\documentstyle[11pt,a4wide]{article}
\include{epsf}
\newcommand{\fig}[4]
   {\begin{figure}[htb]
        \centerline{
                \epsfxsize #2
                \leavevmode\epsffile{#1}}
                \caption{#3}
                \label{#4}
    \end{figure}
   }

\begin{document}

\title{MBT: A Memory-Based Part of Speech Tagger-Generator}

\begin{small}
\author{{\em Walter Daelemans, Jakub Zavrel}\\
Computational Linguistics and AI\\
Tilburg University\\
P.O. Box 90153, NL-5000 LE Tilburg\\
{\tt \{walter.daelemans,zavrel\}@kub.nl}\\
{\em Peter Berck, Steven Gillis}\\
Center for Dutch Language and Speech\\
University of Antwerp\\
Universiteitsplein 1, B-2610 Wilrijk\\
{\tt \{peter.berck,steven.gillis\}@uia.ua.ac.be}}
\end{small}

\date{{\tt In proceedings WVLC 1996, Copenhagen}}

\maketitle 

\abstract{We introduce a memory-based approach to part of speech
tagging.  Memory-based learning is a form of supervised learning based
on similarity-based reasoning. The part of speech tag of a word in a
particular context is extrapolated from the most similar cases held in
memory. Supervised learning approaches are useful when a tagged corpus
is available as an example of the desired output of the tagger.  Based
on such a corpus, the tagger-generator automatically builds a tagger
which is able to tag new text the same way, diminishing development
time for the construction of a tagger considerably. Memory-based
tagging shares this advantage with other statistical or machine
learning approaches. Additional advantages specific to a memory-based
approach include (i) the relatively small tagged corpus size
sufficient for training, (ii) incremental learning, (iii) explanation
capabilities, (iv) flexible integration of information in case
representations, (v) its non-parametric nature, (vi) reasonably good
results on unknown words without morphological analysis, and (vii)
fast learning and tagging.  In this paper we show that a large-scale
application of the memory-based approach is feasible: we obtain a
tagging accuracy that is on a par with that of known statistical
approaches, and with attractive space and time complexity properties
when using {\em IGTree}, a tree-based formalism for indexing and
searching huge case bases.} The use of IGTree has as additional
advantage that optimal context size for disambiguation is dynamically
computed.

\section{Introduction}

Part of Speech (POS) tagging is a process in which syntactic
categories are assigned to words.  It can be seen as a mapping from
sentences to strings of tags.

Automatic tagging is useful for a number of applications: as a
preprocessing stage to parsing, in information retrieval, in text to
speech systems, in corpus linguistics, etc.  The two factors
determining the syntactic category of a word are its lexical
probability (e.g. without context, {\em man} is more probably a noun
than a verb), and its contextual probability (e.g.  after a pronoun,
{\em man} is more probably a verb than a noun, as in {\em they man the
boats}).  Several approaches have been proposed to construct automatic
taggers.  Most work on statistical methods has used n-gram models or
Hidden Markov Model-based taggers (e.g. Church, 1988; DeRose, 1988;
Cutting et al. 1992; Merialdo, 1994, etc.). In these approaches, a tag
sequence is chosen for a sentence that maximizes the product of
lexical and contextual probabilities as estimated from a tagged
corpus.

In rule-based approaches, words are assigned a tag based on a set of
rules and a lexicon. These rules can either be hand-crafted (Garside
et al., 1987; Klein \& Simmons, 1963; Green \& Rubin, 1971), or
learned, as in Hindle (1989) or the transformation-based error-driven
approach of Brill (1992).

In a memory-based approach, a set of cases is kept in memory. Each
case consists of a word (or a lexical representation for the word)
with preceding and following context, and the corresponding category
for that word in that context.  A new sentence is tagged by selecting
for each word in the sentence and its context the most similar case(s)
in memory, and extrapolating the category of the word from these
`nearest neighbors'. A memory-based approach has features of both
learning rule-based taggers (each case can be regarded as a very
specific rule, the similarity based reasoning as a form of conflict
resolution and rule selection mechanism) and of stochastic taggers: it
is fundamentally a form of k-nearest neighbors (k-nn) modeling, a
well-known non-parametric statistical pattern recognition
technique. The approach in its basic form is computationally
expensive, however; each new word in context that has to be tagged,
has to be compared to each pattern kept in memory. In this paper we
show that a heuristic case base compression formalism (Daelemans et
al., 1996), makes the memory-based approach computationally
attractive.

\section{Memory-Based Learning}

Memory-based Learning is a form of {\em supervised, inductive learning
from examples}.  Examples are represented as a vector of feature
values with an associated category label.  During training, a set of
examples (the training set) is presented in an incremental fashion to
the classifier, and added to memory.  During testing, a set of
previously unseen feature-value patterns (the test set) is presented
to the system. For each test pattern, its distance to all examples in
memory is computed, and the category of the least distant instance(s)
is used as the predicted category for the test pattern. The approach
is based on the assumption that reasoning is based on direct reuse of
stored experiences rather than on the application of knowledge (such
as rules or decision trees) abstracted from experience.

In AI, the concept has appeared in several disciplines (from computer
vision to robotics), using terminology such as similarity-based,
example-based, memory-based, exemplar-based, case-based, analogical,
lazy, nearest-neighbour, and instance-based (Stanfill and Waltz, 1986;
Kolodner, 1993; Aha et al.  1991; Salzberg, 1990). Ideas about this
type of analogical reasoning can be found also in non-mainstream
linguistics and pyscholinguistics (Skousen, 1989; Derwing \& Skousen,
1989; Chandler, 1992; Scha, 1992).  In computational linguistics
(apart from incidental computational work of the linguists referred to
earlier), the general approach has only recently gained some
popularity: e.g., Cardie (1994, syntactic and semantic
disambiguation); Daelemans (1995, an overview of work in the early
nineties on memory-based computational phonology and morphology);
Jones (1996, an overview of example-based machine translation
research); Federici and Pirrelli (1996).

\subsection{Similarity Metric}

Performance of a memory-based system (accuracy on the test set)
crucially depends on the distance metric (or similarity metric) used.
The most straightforward distance metric would be the one in equation
(1), where $X$ and $Y$ are the patterns to be compared, and $\delta
(x_{i},y_{i})$ is the distance between the values of the $i$-th
feature in a pattern with $n$ features.

\begin{equation}
\Delta(X,Y) = \sum_{i=1}^{n} \delta(x_{i},y_{i})
\end{equation}

Distance between two values is measured using equation (2), an overlap
metric, for symbolic features (we will have no numeric features in the
tagging application). 

\begin{equation}
\delta(x_{i}, y_{i}) = 0\ if\ x_{i} = y_{i},\ else\ 1
\end{equation}

We will refer to this approach as IB1 (Aha et al., 1991). We extended
the algorithm described there in the following way: in case a pattern
is associated with more than one category in the training set
(i.e. the pattern is ambiguous), the distribution of patterns over the
different categories is kept, and the most frequently occurring
category is selected when the ambiguous pattern is used to extrapolate
from.

In this distance metric, all features describing an example are
interpreted as being equally important in solving the classification
problem, but this is not necessarily the case.  In tagging, the focus
word to be assigned a category is obviously more relevant than any of
the words in its context.  We therefore weigh each feature with its
{\em information gain}; a number expressing the average amount of
reduction of training set information entropy when knowing the value
of the feature (Daelemans \& van de Bosch, 1992, Quinlan, 1993; Hunt
et al. 1966) (Equation 3). We will call this algorithm IB-IG.


\begin{equation}
\Delta(X,Y) = \sum_{i=1}^{n} G(f_{i}) \delta(x_{i},y_{i})
\end{equation}

\section{IGTrees}

Memory-based learning is an expensive algorithm: of each test item,
all feature values must be compared to the corresponding feature
values of all training items. Without optimisation, it has an
asymptotic retrieval complexity of $O(NF)$ (where $N$ is the number of
items in memory, and $F$ the number of features). The same asymptotic
complexity is of course found for memory storage in this approach.  We
use IGTrees (Daelemans et al. 1996) to compress the memory. IGTree is
a heuristic approximation of the IB-IG algorithm. 


\subsection{The IGTree Algorithms}

IGTree combines two algorithms: one for compressing a case base into a
trees, and one for retrieving classification information from these
trees.  During the construction of IGTree decision trees, cases are
stored as paths of connected nodes.  All nodes contain a test (based
on one of the features) and a class label (representing the default
class at that node). Nodes are connected via arcs denoting the
outcomes for the test (feature values). A feature relevance ordering
technique (in this case information gain, see Section 2.1) is used to
determine the order in which features are used as tests in the
tree. This order is fixed in advance, so the maximal depth of the tree
is always equal to the number of features, and at the same level of
the tree, all nodes have the same test (they are an instance of {\em
oblivious decision trees}; cf. Langley \& Sage, 1994). The reasoning
behind this reorganisation (which is in fact a compression) is that
when the computation of feature relevance points to one feature
clearly being the most important in classification, search can be
restricted to matching a test case to those stored cases that have the
same feature value at that feature. Besides restricting search to
those memory cases that match only on this feature, the case memory
can be optimised by further restricting search to the second most
important feature, followed by the third most important feature,
etc. A considerable compression is obtained as similar cases share
partial paths.

Instead of converting the case base to a tree in which all cases are
fully represented as paths, storing all feature values, we compress
the tree even more by restricting the paths to those input feature
values that disambiguate the classification from all other cases in
the training material. The idea is that it is not necessary to fully
store a case as a path when only a few feature values of the case make
its classification unique. This implies that feature values that do
not contribute to the disambiguation of the case classification (i.e.,
the values of the features with lower feature relevance values than
the the lowest value of the disambiguating features) are {\em not}\/
stored in the tree. In our tagging application, this means that only
context feature values that actually contribute to disambiguation are
used in the construction of the tree. 

Leaf nodes contain the unique class label corresponding to a path in
the tree.  Non-terminal nodes contain information about the {\em most
probable}\/ or {\em default}\/ classification given the path thus far,
according to the bookkeeping information on class occurrences
maintained by the tree construction algorithm. This extra information
is essential when using the tree for classification.  Finding the
classification of a new case involves traversing the tree (i.e.,
matching all feature values of the test case with arcs in the order of
the overall feature information gain), and either retrieving a
classification when a leaf is reached, or using the default
classification on the last matching non-terminal node if a
feature-value match fails.

A final compression is obtained by pruning the derived tree.  All
leaf-node daughters of a mother node that have the same class as that
node are removed from the tree, as their class information does not
contradict the default class information already present at the mother
node. Again, this compression does not affect IGTree's generalisation
performance.

The recursive algorithms for tree construction (except the final
pruning) and retrieval are given in Figures \ref {build-igtree} and
\ref{search-igtree}. For a detailed discussion, see Daelemans et
al. (1996).

\begin{figure}[t]
{\footnotesize
\rule[2mm]{\textwidth}{1.0mm}
\begin{description}
\item
Procedure {\bf BUILD-IG-TREE}:
\item
Input:
\begin{itemize}
\item A training set $T$ of cases with their classes (start
value: a full case base),
\item an information-gain-ordered list of features (tests)
$F_{i}...F_{n}$ (start value: $F_{1}...F_{n}$).
\end{itemize}
\item
Output: A (sub)tree.
\begin{enumerate}
\item
If $T$ is unambiguous (all cases in $T$ have the same class $c$),
create a leaf node with class label $c$.  
\item
Else if $i = (n+1) $, create a leaf node with as label the class occurring
most frequently in $T$.
\item
Otherwise, until $i=n$ (the number of features)
\begin{itemize}
\item
Select the first feature (test) $F_{i}$ in $F_{i}...F_{n}$, and
construct a new node $N$ for feature $F_{i}$, and as default class
$c$ (the class occurring most frequently in $T$).
\item
Partition $T$ into subsets $T_{1}...T_{m}$ according to the values
$v_{1}...v_{m}$ which occur for $F_{i}$ in $T$ (cases with the
same value for this feature in the same subset).
\item
For each $j \epsilon \{ 1,...,m \}$: BUILD-IG-TREE $( T_{j},
F_{i+1}...F_{n} )$, \\ 
connect the root of this subtree to $N$ and label the arc with $v_{j}$.
\end{itemize}
\end{enumerate}
\end{description}
\rule[2mm]{\textwidth}{1.0mm}
}
\caption{Algorithm for building IGTrees (`BUILD-IG-TREE').}\label{build-igtree}
\end{figure}

\begin{figure}[htb]
{\footnotesize
\rule[2mm]{\textwidth}{1.0mm}
\begin{description}
\item
Procedure {\bf SEARCH-IG-TREE}:
\item
Input:
\begin{itemize}
\item The root node $N$ of a subtree (start value: top node of a complete
IGTree),
\item an unlabeled case $I$ with information-gain-ordered feature
values $f_{i} ... f_{n}$ (start value: $f_{1} ... f_{n}$).
\end{itemize}
\item
Output: A class label. 
\begin{enumerate}
\item
If $N$ is a leaf node, output default class $c$
associated with this node.
\item
Otherwise, if test $F_{i}$ of the current node does not originate an
arc labeled with $f_{i}$, output default class
$c$ associated with $N$.
\item
Otherwise,
\begin{itemize}
\item new node $M$ is the end node of the arc originating from $N$
with as label $f_{i}$.
\item
SEARCH-IG-TREE $( M , f_{i+1} ... f_{n} )$
\end{itemize}
\end{enumerate}
\end{description}
\rule[2mm]{\textwidth}{1.0mm}
}
\caption{Algorithm for searching IGTrees
(`SEARCH-IG-TREE').}\label{search-igtree}
\end{figure}

\subsection{IGTree Complexity}

The asymptotic complexity of IGTree (i.e, in the worst case) is
extremely favorable.  Complexity of searching a query pattern in the
tree is proportional to $F*log(V)$, where $F$ is the number of
features (equal to the maximal depth of the tree), and $V$ is the
average number of values per feature (i.e., the average branching
factor in the tree). In IB1, search complexity is $O(N*F)$ (with $N$
the number of stored cases).  Retrieval by search in the tree is
independent from the number of training cases, and therefore
especially useful for large case bases.  Storage requirements are
proportional to $N$ (compare $O(N*F)$ for IB1).  Finally, the cost of
building the tree on the basis of a set of cases is proportional to
$N*log(V)*F$ in the worst case (compare $O(N)$ for training in IB1).

In practice, for our part-of-speech tagging experiments, IGTree
retrieval is 100 to 200 times faster than normal memory-based
retrieval, and uses over 95\% less memory.

\section{Architecture of the Tagger}

The architecture takes the form of a {\em tagger generator}: given a
corpus tagged with the desired tag set, a POS tagger is generated
which maps the words of new text to tags in this tag set according to
the same systematicity. The construction of a POS tagger for a
specific corpus is achieved in the following way. Given an annotated
corpus, three datastructures are automatically extracted: a {\em
lexicon}, a case base for {\em known words} (words occurring in the
lexicon), and a case base for {\em unknown words}. Case Bases are
indexed using IGTree. During tagging, each word in the text to be
tagged is looked up in the lexicon. If it is found, its lexical
representation is retrieved and its context is determined, and the
resulting pattern is looked up in the known words case base. When a
word is not found in the lexicon, its lexical representation is
computed on the basis of its form, its context is determined, and the
resulting pattern is looked up in the unknown words case base. In each
case, output is a best guess of the category for the word in its
current context. In the remainder of this section, we will describe
each step in more detail. We start from a {\em training set} of tagged
sentences $T$.

\subsection{Lexicon Construction}

A lexicon is extracted from $T$ by computing for each word in $T$ the
number of times it occurs with each category. E.g. when using the
first 2 million words of the Wall Street Journal corpus\footnote{ACL
Data Collection Initiative CD-ROM 1, September 1991.} as $T$, the word
{\em once} would get the lexical definition {\em RB: 330; IN: 77},
i.e. {\em once} was tagged 330 times as an adverb, and 77 times as a
preposition/subordinating conjunction.\footnote{We disregarded a
category associated with a word when less than 10\% of the word tokens
were tagged with that category. This way, noise in the training
material is filtered out. The value for this parameter will have to be
adapted for other training sets, and was chosen here to maximise
generalization accuracy (accuracy on tagging unseen text).}

Using these lexical definitions, a new, possibly ambiguous, tag is
produced for each word type.  E.g. {\em once} would get a new tag,
representing the category of words which can be both adverbs and
prepositions/conjunctions (RB-IN).  Frequency order is taken into
account in this process: if there would be words which, like {\em
once}, can be RB or IN, but more frequently IN than RB (e.g. the word
{\em below}), then a different tag (IN-RB) is assigned to these words.
The original tag set, consisting of 44 morphosyntactic tags, was
expanded this way to 419 (possibly ambiguous) tags. In the WSJ
example, the resulting lexicon contains 57962 word types, 7464 (13\%)
of which are ambiguous. On the same training set, 76\% of word {\em
tokens} are ambiguous.

When tagging a new sentence, words are looked up in the
lexicon. Depending on whether or not they can be found there, a case
representation is constructed for them, and they are retrieved from
either the known words case base or the unknown words case base.

\subsection{Known Words}

A windowing approach (Sejnowski \& Rosenberg, 1987) was used to
represent the tagging task as a classification problem.  A case
consists of information about a focus word to be tagged, its left and
right context, and an associated category (tag) valid for the focus
word in that context.

There are several types of information which can be stored in the case
base for each word, ranging from the words themselves to intricate
lexical representations. In the preliminary experiments described in
this paper, we limited this information to the possibly ambiguous tags
of words (retrieved from the lexicon) for the focus word and its
context to the right, and the disambiguated tags of words for the left
context (as the result of earlier tagging decisions).  Table
\ref{known-case-rep} is a sample of the case base for the first
sentence of the corpus ({\em Pierre Vinken, 61 years old, will join
the board as a nonexecutive director nov. 29}) when using this case
representation. The final column shows the target category; the
disambiguated tag for the focus word. We will refer to this case
representation as {\tt ddfat} (d for disambiguated, f for focus, a for
ambiguous, and t for target). The information gain values are given as
well.

{\small
\begin{table}[htb]
\caption{Case representation and information gain pattern for known
words.}\label{known-case-rep}
\begin{center}
\begin{tabular}{|l|llll|l|}\hline
word & \multicolumn{5}{|c|}{case representation}\\
 & d & d & f & a & t \\\hline
IG & .06 & .22 & .82 & .23 & \\
\hline
Pierre & = & = & np & np & np\\
Vinken & = & np & np & ,  & np\\
, & np & np & ,  & cd & ,\\
61 & np & ,  & cd & nns & cd\\
years & , & cd & nns & jj-np & nns\\
old & cd & nns & jj-np & , & jj\\
\hline
\end{tabular}
\end{center}
\end{table}
} 

A search among a selection of different context sizes suggested {\tt
ddfat} as a suitable case representation for tagging known words. An
interesting property of memory-based learning is that case
representations can be easily extended with different sources of
information if available (e.g. feedback from a parser in which the
tagger operates, semantic types, the words themselves, lexical
representations of words obtained from a different source than the
corpus, etc.). The information gain feature relevance ordering
technique achieves a delicate relevance weighting of different
information sources when they are fused in a single case
representation. The window size used by the algorithm will also
dynamically change depending on the information present in the context
for the disambiguation of a particular focus symbol (see Sch\"utze et
al., 1994, and Pereira et al., 1995 for similar approaches).

\subsection{Unknown Words}

If a word is not present in the lexicon, its ambiguous category cannot
be retrieved. In that case, a category can be guessed only on the
basis of the {\em form} or the {\em context} of the word. Again, we
take advantage of the data fusion capabilities of a memory-based
approach by combining these two sources of information in the case
representation, and having the information gain feature relevance
weighting technique figure out their relative relevance (see Schmid,
1994; Samuelsson, 1994 for similar solutions).

In most taggers, some form of morphological analysis is performed on
unknown words, in an attempt to relate the unknown word to a known
combination of known morphemes, thereby allowing its association with
one or more possible categories. After determining this ambiguous
category, the word is disambiguated using context knowledge, the same
way as known words.  Morphological analysis presupposes the
availability of highly language-specific resources such as a morpheme
lexicon, spelling rules, morphological rules, and heuristics to
prioritise possible analyses of a word according to their
plausibility. This is a serious knowledge engineering bottleneck when
the goal is to develop a language and annotation-independent tagger
generator.

In our memory-based approach, we provide morphological information
(especially about suffixes) indirectly to the tagger by encoding the
three last letters of the word as separate features in the case
representation. The first letter is encoded as well because it
contains information about prefix and capitalization of the
word. Context information is added to the case representation in a
similar way as with known words. It turned out that in combination
with the `morphological' features, a context of one disambiguated tag
of the word to the left of the unknown word and one ambiguous category
of the word to the right, gives good results. We will call this case
representation {\tt pdassst}:\footnote{These parameters (optimal
context size and number of suffix features) were again optimised for
generalization accuracy.}  three suffix letters (s), one prefix letter
(p), one left disambiguated context words (d), and one ambiguous right
context word (a). As the chance of an unknown word being a function
word is small, and cases representing function words may interfere
with correct classification of open-class words, only open-class words
are used during construction of the unknown words case base.

Table \ref{unknown-case-rep} shows part of the case base for unknown
words.

{\small
\begin{table}[htb]
\caption{Case representation and information gain pattern for unknown
words.}\label{unknown-case-rep}
\begin{center}
\begin{tabular}{|l|llllll|l|}\hline
word & \multicolumn{7}{|c|}{case representation}\\
 & p & d & a & s & s & s & t \\\hline
IG & .21 & .21 & .14 & .15 & .20 & .32 & \\
\hline
Pierre & P & = & np & r & r & e & np\\
Vinken & V & np & ,  & k & e & n & np\\
61 & 6 & ,  & nns & = & 6 & 1 & cd\\
years & y & cd & jj-np & a & r & s & nns\\
old & o & nns & , & o & l & d & jj\\
\hline
\end{tabular}
\end{center}
\end{table}
}

\subsection{Control}

Figure \ref{architecture} shows the architecture of the
tagger-generator: a tagger is produced by extracting a lexicon and two
case-bases from the tagged example corpus. During tagging, the control
is the following: words are looked up in the lexicon and separated
into known and unknown words. They are retrieved from the known words
case base and the unknown words case base, respectively. In both
cases, context is used, in the case of unknown words, the first and
three last letters of the word are used instead of the ambiguous tag
for the focus word. As far as {\em disambiguated} tags for left
context words are used, these are of course not obtained by retrieval
from the lexicon (which provides ambiguous categories), but by using
the previous decisions of the tagger.

\fig{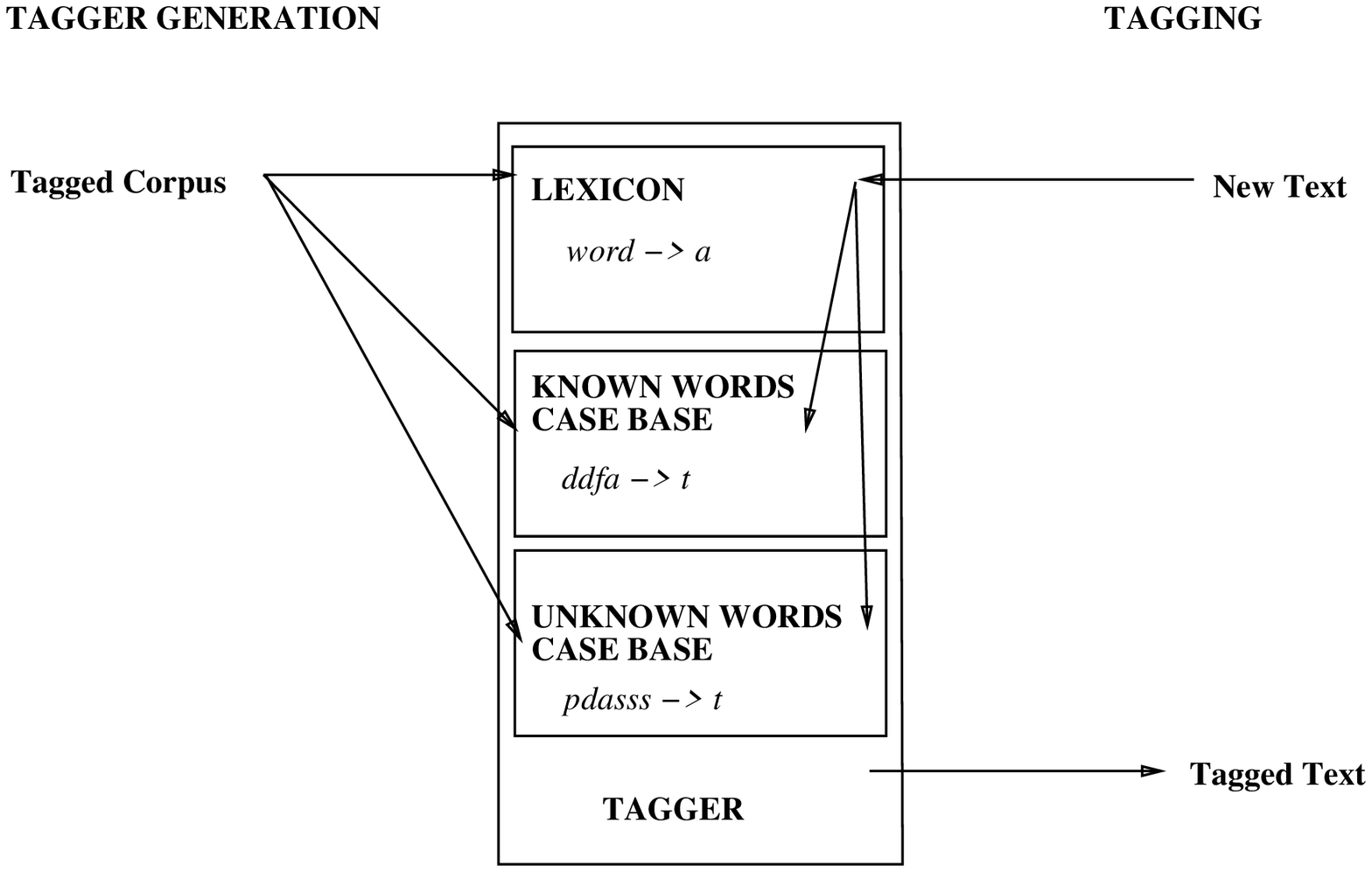}{10cm}{Architecture of the tagger-generator: flow of control.}{architecture}

\subsection{IGTrees for Tagging}

As explained earlier, both case bases are implemented as IGTrees. For
the known words case base, paths in the tree represent {\em variable
size} context widths. The first feature (the expansion of the root
node of the tree) is the focus word, then context features are added
as further expansions of the tree until the context disambiguates the
focus word completely. Further expansion is halted at that point.  In
some cases, short context sizes (corresponding to bigrams, e.g.) are
sufficient to disambiguate a focus word, in other cases, more context
is needed. IGTrees provide an elegant way of automatic determination
of optimal context size. In the unknown words case base, the trie
representation provides an automatic integration of information about
the form and the context of a focus word not encountered before. In
general, the top levels of the tree represent the morphological
information (the three suffix letter features and the prefix letter),
while the deeper levels contribute contextual disambiguation.

\section{Experiments}

In this section, we report first results on our memory-based tagging
approach. In a first set of experiments, we compared our IGTree
implementation of memory-based learning to more traditional
implementations of the approach. In further experiments we studied the
performance of our system on predicting the category of both known and
unknown words.

\subsection*{Experimental Set-up}

The experimental methodology was taken from Machine Learning practice
(e.g. Weiss \& Kulikowski, 1991): independent training and test sets
were selected from the original corpus, the system was trained on the
training set, and the generalization accuracy (percentage of correct
category assignments) was computed on the independent test
set. Storage and time requirements were computed as well. Where
possible, we used a 10-fold cross-validation approach. In this
experimental method, a data set is partitioned ten times into 90\%
training material, and 10\% testing material. Average accuracy
provides a reliable estimate of the generalization accuracy.

\subsection{Experiment 1: Comparison of  Algorithms}

Our goal is to adhere to the concept of memory-based learning with
full memory while at the same time keeping memory and processing speed
within attractive bounds. To this end, we applied the IGTree formalism
to the task. In order to prove that IGTree is a suitable candidate for
practical memory-based tagging, we compared three memory-based
learning algorithms: (i) IB1, a slight extension (to cope with
symbolic values and ambiguous training items) of the well-known k-nn
algorithm in statistical pattern recognition (see Aha et al., 1991),
(ii) IB1-IG, an extension of IB1 which uses feature relevance
weighting (described in Section 2), and (iii) IGTree, a memory- and
processing time saving heuristic implementation of IB1-IG (see Section
3). Table \ref{exp1} lists the results in generalization accuracy,
storage requirements and speed for the three algorithms using a {\tt
ddfat} pattern, a 100,000 word training set, and a 10,000 word test
set. In this experiment, accuracy was tested on known words only.

\begin{table}[htb]
\caption{Comparison of three memory-based learning
techniques.}\label{exp1}
\begin{center}
\begin{tabular}{|l|r|r|r|}\hline
Algorithm & Accuracy & Time & Memory (Kb) \\
\hline
IB1 & 92.5 & 0:43:34 & 977 \\
IB1-IG & 96.0 & 0:49:45 & 977 \\
IGTree & 96.0 & 0:00:29 & 35 \\
\hline
\end{tabular}
\end{center}
\end{table}

The IGTree version turns out to be better or equally good in terms of
generalization accuracy, but also is more than 100 times faster for
tagging of new words\footnote{In training, i.e. building the case
base, IB1 and IB1-IG (4 seconds) are faster than IGTree (26 seconds)
because the latter has to build a tree instead of just storing the
patterns.}, and compresses the original case base to 4\% of the size
of the original case base. This experiment shows that for this
problem, we can use IGTree as a time and memory saving approximation
of memory-based learning (IB-IG version), without loss in
generalization accuracy.  The time and speed advantage of IGTree grows
with larger training sets.

\subsection{Experiment 2: Learning Curve}

A ten-fold cross-validation experiment on the first two million words
of the WSJ corpus shows an average generalization performance of
IGTree (on known words only) of 96.3\%. We did 10-fold
cross-validation experiments for several sizes of datasets (in steps
of 100,000 memory items), revealing the learning curve in Figure
\ref{curve}.  Training set size is on the X-axis, generalization
performance as measured in a 10-fold cross-validation experiment is on
the Y-axis.  the `error' range indicate averages plus and minus one
standard deviation on each 10-fold cross-validation
experiment.\footnote{We are not convinced that variation in the
results of the experiments in a 10-fold-cv set-up is statistically
meaningful (the 10 experiments are not independent), but follow common
practice here.}

\fig{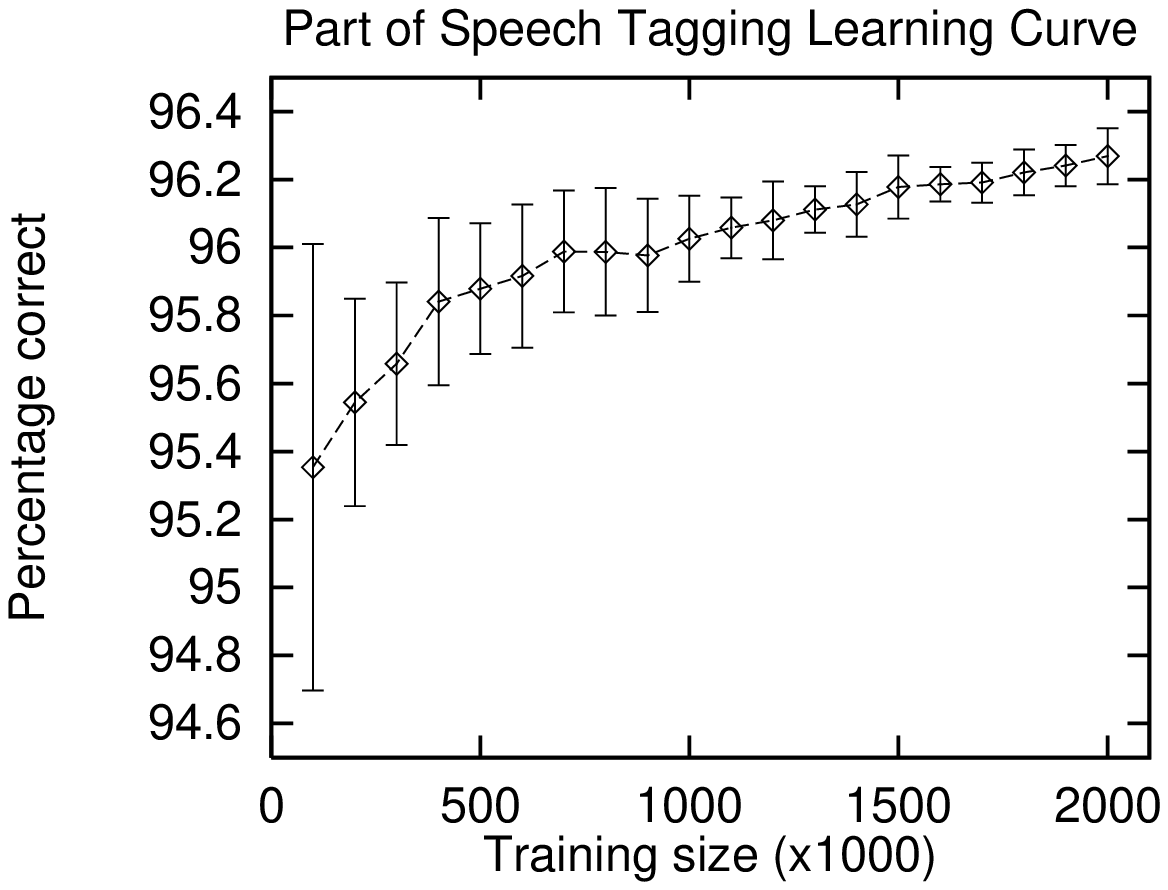}{8cm}{Learning curve for tagging.}{curve}

Already at small data set sizes, performance is relatively high. With
increasingly larger data sets, the performance becomes more stable
(witness the error ranges). It should be noted that in this
experiment, we assumed correctly disambiguated tags in the left
context. In practice, when using our tagger, this is of course not the
case because the disambiguated tags in the left context of the current
word to be tagged are the result of a previous decision of the tagger,
which may be a mistake. To test the influence of this effect we
performed a third experiment.

\subsection{Experiment 3: Overall Accuracy}

We performed the complete tagger generation process on a 2 million
words training set (lexicon construction and known and unknown words
case-base construction), and tested on 200,000 test words. Performance
on known words, unknown words, and total are given in Table
\ref{exp3}. In this experiment, numbers were not stored in the known
words case base; they are looked up in the unknown words case base.

\begin{table}[htb]
\caption{Accuracy of IGTree tagging on known and unknown words}\label{exp3}
\begin{center}
\begin{tabular}{|r|r|r|}\hline
     & Accuracy & Percentage \\
\hline
Known & 96.7 & 94.5 \\
Unknown & 90.6 & 5.5 \\
Total & 96.4 & 100.0 \\
\hline
\end{tabular}
\end{center}
\end{table}

\section{Related Research}

A case-based approach, similar to our memory-based approach, was also
proposed by Cardie (1993a, 1994) for sentence analysis in limited
domains (not only POS tagging but also semantic tagging and structural
disambiguation). We will discuss only the reported POS tagging results
here. Using a fairly complex case representation based on output from
the CIRCUS conceptual sentence analyzer (22 local context features
describing syntactic and semantic information about a five-word window
centered on the word to be tagged, including the words themselves, and
11 global context features providing information about the major
constituents parsed already), and with a tag set of 18 tags (7
open-class, 11 closed class), she reports a 95\% tagging accuracy. A
decision-tree learning approach to feature selection is used in this
experiment (Cardie, 1993b, 1994) to discard irrelevant features.
Results are based on experiments with 120 randomly chosen sentences
from the TIPSTER JV corpus (representing 2056 cases).  Cardie (p.c.)
reports 89.1\% correct tagging for unknown words. Percentage unknown
words was 20.6\% of the test words, and overall tagging accuracy
(known and unknown) 95\%. Notice that her algorithm gives no initial
preference to training cases that match the test word during its
initial case retrieval. On the other hand, after retrieving the top k
cases, the algorithm does prefer those cases that match the test word
when making its final predictions.  So, it's understandable that the
algorithm is doing better on words that it's seen during training as
opposed to unknown words.

In our memory-based approach, feature weighting (rather than feature
selection) for determining the relevance of features is integrated
more smoothly with the similarity metric, and our results are based on
experiments with a larger corpus (3 million cases). Our case
representation is (at this point) simpler: only the (ambiguous) tags,
not the words themselves or any other information are used. The most
important improvement is the use of IGTree to index and search the
case base, solving the computational complexity problems a case-based
approach would run into when using large case bases.

An approach based on k-nn methods (such as memory-based and case-based
methods) is a statistical approach, but it uses a different kind of
statistics than Markov model-based approaches.  K-nn is a
non-parametric technique; it assumes no fixed type of distribution of
the data.  The most important advantages compared to current
stochastic approaches are that (i) few training items (a small tagged
corpus) are needed for relatively good performance, (ii) the approach
is incremental: adding new cases does not require any recomputation of
probabilities, and (iii) it provides explanation capabilities, and
(iv) it requires no additional smoothing techniques to avoid
zero-probabilities; the IGTree takes care of that.

Compared to hand-crafted rule-based approaches, our approach provides
a solution to the knowledge-acquisition and reusability bottlenecks,
and to robustness and coverage problems (similar advantages motivated
Markov model-based statistical approaches). Compared to {\em learning}
rule-based approaches such as the one by Brill (1992), a k-nn approach
provides a uniform approach for all disambiguation tasks, more
flexibility in the engineering of case representations, and a more
elegant approach to handling of unknown words (see e.g. Cardie 1994).

\section{Conclusion}

We have shown that a memory-based approach to large-scale tagging is
feasible both in terms of accuracy (comparable to other statistical
approaches), and also in terms of computational efficiency (time and
space requirements) when using IGTree to compress and index the case
base. The approach combines some of the best features of learned
rule-based and statistical systems (small training corpora needed,
incremental learning, understandable and explainable behavior of the
system). More specifically, memory-based tagging with IGTrees has the
following advantages.

\begin{itemize}
\item
Accurate generalization from small tagged corpora. Already at small
corpus size (300--400 K tagged words), performance is good. These
corpus sizes can be easily handled by our system.
\item
Incremental learning. New `cases' (e.g. interactively corrected output
of the tagger) can be incrementally added to the case bases,
continually improving the performance of the overall system.
\item
Explanation capabilities. To explain the classification behavior of
the system, a path in the IGTree (with associated defaults) can be
provided as an explanation, as well as nearest neighbors from which
the decision was extrapolated.
\item
Flexible integration of information sources. The feature weighting
method takes care of the optimal fusing of different sources of
information (e.g. word form and context), automatically.
\item
Automatic selection of optimal context. The IGTree mechanism (when
applied to the known words case base) automatically decides on the
optimal context size for disambiguation of focus words.
\item
Non-parametric estimation. The IGTree formalism provides automatic,
non\-parametric estimation of classifications for low-frequency
contexts (it is similar in this respect to backed-off training), but
avoids non-optimal estimation due to false intuitions or
non-convergence of the gradient-descent procedure used in some
versions of backed-off training.
\item
Reasonably good results on unknown words without morphological
analysis. On the WSJ corpus, unknown words can be predicted (using
context and word form information) for more than 90\%.
\item
Fast learning and tagging. Due to the favorable complexity properties
of IGTrees (lookup time in IGTrees is independent on number of cases),
both tagger generation and tagging are extremely fast. Tagging speed
in our current implementation is about 1000 words per second.
\end{itemize}

We have barely begun to optimise the approach: a more intelligent
similarity metric would also take into account the differences in
similarity between different values of the same feature. E.g. the
similarity between the tags rb-in-nn and rb-in should be bigger than
the similarity between rb-in and vb-nn. Apart from linguistic
engineering refinements of the similarity metric, we are currently
experimenting with statistical measures to compute such more
fine-grained similarities (e.g. Stanfill \& Waltz, 1986, Cost \&
Salzberg, 1994).

\section*{Acknowledgements}

Research of the first author was done while he was a visiting scholar
at NIAS (Netherlands Institute for Advanced Studies) in Wassenaar.
Thanks to Antal van den Bosch, Ton Weijters, and Gert Durieux for
discussions about tagging, IGTree, and machine learning of natural
language.

{\small
\section*{References}
\begin{description}
\item
Aha,~D.~W., Kibler,~D., \& Albert,~M. (1991).  `Instance-based
learning algorithms'.  {\em Machine Learning}, 7, 37--66.
\item
Brill, E. (1992) `A simple rule-based part-of-speech tagger'.  {\em
Proceedings Third ACL Applied}, Trento, Italy, 152--155.
\item
Cardie, C. (1993a).  `A case-based approach to knowledge acquisition
for domain-specific sentence analysis'.  In {\em AAAI-93}, 798--803.
\item
Cardie, C. (1993b).  `Using Decision Trees to Improve Case-Based
Learning'.  In {\em Proceedings of the Tenth International Conference
on Machine Learning}, 25--32.
\item
Cardie, C. (1994).  `Domain-Specific Knowledge Acquisition for
Conceptual Sentence Analysis'.  Ph.D. Thesis, University of
Massachusetts, Amherst, MA.
\item
Chandler, S. (1992).  `Are rules and modules really necessary for
explaining language?'  {\em Journal of Psycholinguistic research},
22(6): 593--606.
\item
Church, K.  (1988).  `A stochastic parts program and noun phrase
parser for unrestricted text'.  {\em Proceedings Second ACL Applied
NLP}, Austin, Texas, 136--143.
\item
Cost, S. and Salzberg, S.  (1993).  `A weighted nearest neighbour
algorithm for learning with symbolic features.'  {\em Machine
Learning}, 10, 57--78.
\item
Cutting, D., Kupiec, J., Pederson, J., Sibun, P. (1992).  A practical
part of speech tagger.  {\em Proceedings Third ACL Applied NLP},
Trento, Italy, 133--140.
\item
Daelemans, W. (1995).
`Memory-based lexical acquisition and processing.'
In Steffens, P., editor, {\em Machine Translation and the Lexicon},
Lecture Notes in Artificial Intelligence 898. Berlin: Springer,
85--98. 
\item
Daelemans,~W., Van~den~Bosch,~A. (1992).  `Generalisation performance
of backpropagation learning on a syllabification task.'  In
M. Drossaers \& A. Nijholt (Eds.), {\em TWLT3: Connectionism and
Natural Language Processing}. Enschede: Twente University, 27--38.
\item
Daelemans, W., Van den Bosch, A., Weijters, T. (1996).  `IGTree: Using
Trees for Compression and Classification in Lazy Learning Algorithms.'
In Aha, D. (ed.). {\em AI Review Special Issue on Lazy Learning},
forthcoming.
\item
DeRose, S. (1988).  `Grammatical category disambiguation by
statistical optimization. ' {\em Computational Linguistics} 14,
31--39.
\item
Derwing, B. L. and Skousen, R. (1989).  `Real Time Morphology:
Symbolic Rules or Analogical Networks'.  {\em Berkeley Linguistic
Society} 15: 48--62.
\item
Federici S. and V. Pirelli. (1996). `Analogy, Computation and
Linguistic Theory.' In Jones, D. (ed.) {\em New Methods in Language
Processing}. London: UCL Press, forthcoming.
\item
Garside, R., Leech, G. and Sampson, G. (1987).  {\em The computational
analysis of English: A corpus-based approach}, London: Longman, 1987.
\item
Greene, B.B. and Rubin, G.M. (1971). {\em Automatic Grammatical
Tagging of English}.  Providence RI: Department of Linguistics, Brown
University.
\item
Hindle, Donald. (1989). `Acquiring disambiguation rules from text.'
In {\em Proceedings, 27th Annual Meeting of the Association for
Computational Linguistics}, Vancouver, BC.
\item 
Hunt, E., J. Marin, P. Stone. (1966). {\em Experiments in Induction.}
New York: Academic Press.
\item
Jones, D. {\em Analogical Natural Language Processing}. London: UCL
Press, 1996.
\item
Klein S. and Simmons, R. (1963).  `A grammatical approach to
grammatical coding of English words.'  {\em JACM} 10, 334--347.
\item
Kolodner, J. (1993).  {\em Case-Based Reasoning.}  San Mateo: Morgan
Kaufmann.
\item
Langley,~P. and Sage,~S. (1994).  `Oblivious decision trees and
abstract cases.'  In D.~W.~Aha (Ed.), {\em Case-Based Reasoning:
Papers from the 1994 Workshop} (Technical Report WS-94-01).  Menlo
Park, CA: AAAI Press.
\item
Merialdo, B. ( 1994).  `Tagging English Text with a Probabilistic
Model.'  {\em Computational Linguistics} 20 (2), 155--172.
\item
Pereira, F., Y. Singer, N. Tishby. (1995). `Beyond Word N-grams.' {\em
Proceedings Third Workshop on Very Large Corpora}, MIT, Cambridge
Mass., 95--106.
\item
Quinlan, J. (1993).  {\em C4.5: Programs for Machine Learning}.  San
Mateo, CA: Morgan Kaufmann.
\item
Salzberg, S. (1990) `A nearest hyperrectangle learning method'.  {\em
Machine Learning} 6, 251--276.
\item
Samuelsson, C. (1994) `Morphological Tagging Based Entirely on
Bayesian Inference.' In {\em Proceedings of the 9th Nordic Conference
on Computational Linguistics}, Stockholm University, Sweden, 1994.
\item
Scha, R. (1992) `Virtuele Grammatica's en Creatieve Algoritmen.'  {\em
Gramma/TTT} 1 (1), 57--77.
\item
Schmid, H. (1994) `Part-of-speech tagging with neural networks.' In
{\em Proceedings of COLING}, Kyoto, Japan.
\item
Sch\"utze, H., and Y. Singer. (1994) `Part-of-speech Tagging Using a
Variable Context Markov Model' {\em Proceedings of ACL 1994}, Las
Cruces, New Mexico.
\item
Skousen, R. (1989).  {\em Analogical Modeling of Language}.
Dordrecht: Kluwer.
\item
Sejnowski,~T.~J., Rosenberg,~C.~S. (1987).  Parallel networks that
learn to pronounce English text.  {\em Complex Systems}, 1, 145--168.
\item
Stanfill, C. and Waltz, D. (1986).  `Toward memory-based reasoning.'
{\em Communications of the ACM}, 29, 1212--1228.
\item
Weiss, S. and Kulikowski, C. (1991).  {\em Computer systems that
learn.}  San-Mateo: Morgan Kaufmann.
\item
\end{description}
}
\end{document}